\documentclass[conference]{IEEEtran}
\IEEEoverridecommandlockouts
\usepackage{cite}
\usepackage{amsmath,amssymb,amsfonts}
\usepackage{tikz}
\usepackage{color}
\usepackage{pgfpages}
\usepackage{pgfplots}
\usepackage{algorithm}
\usepackage{algorithmic}
\usepackage{graphicx}
\usepackage{textcomp}
\usepackage{xcolor}
\usepackage{amsmath}
\usepackage{lipsum}
\usepackage{mathtools}
\usepackage{float}
\usepackage{cuted}
\def\BibTeX{{\rm B\kern-.05em{\sc i\kern-.025em b}\kern-.08em
    T\kern-.1667em\lower.7ex\hbox{E}\kern-.125emX}}
\begin{document}

\title{Quantized Guided Pruning for Efficient Hardware Implementations of Convolutional Neural Networks}
\author{Ghouthi Boukli Hacene$^{1,2}$, Vincent Gripon $^{1,2}$, Matthieu Arzel$^2$,  Nicolas Farrugia$^2$ and Yoshua Bengio$^1$ \\
 $^1$ Universit\'e de Montr\'eal, MILA, $^2$IMT Atlantique, Lab-STICC\\
 (Paper submitted to ISCAS 2019 on October 31, 2018)}

\maketitle

\begin{abstract}

Convolutional Neural Networks (CNNs) are state-of-the-art in numerous computer vision tasks such as object classification and detection. However, the large amount of parameters they contain leads to a high computational complexity and strongly limits their usability in budget-constrained devices such as embedded devices. In this paper, we propose a combination of a new pruning technique and a quantization scheme that effectively reduce the complexity and memory usage of convolutional layers of CNNs, and replace the complex convolutional operation by a low-cost multiplexer. We perform experiments on the CIFAR10, CIFAR100 and SVHN and show that the proposed method achieves almost state-of-the-art accuracy, while drastically reducing the computational and memory footprints. We also propose an efficient hardware architecture to accelerate CNN operations. The proposed hardware architecture is a pipeline and accommodates multiple layers working at the same time to speed up the inference~process. 

\end{abstract}


\section{Introduction and Related Work}
For the past few years, Deep Neural Networks (DNNs), and especially Convolutional Neural Networks (CNNs)~\cite{lecun1998gradient}, have received considerable attention thanks to their remarkable accuracy in computer vision tasks~\cite{iandola2016squeezenet,DBLP:journals/corr/SimonyanZ14a,DBLP:journals/corr/Graham14a,szegedy2015rethinking} such as classification and detection~\cite{lecun2015deep}. However, their need for intensive computations and memory has meant that most of the implementations are based on GPUs, while providing efficient hardware implementations is still a very active and prospective field of research. Therefore, the deployment of CNNs in embedded systems is complex and potentially blocking for many potential~applications. 

To address this limitation, multiple approaches have been proposed in the literature. For example the authors of~\cite{han2015deep} and~\cite{kim2015compression} propose to reduce DNNs' memory footprint by compressing their weights. In these two approaches, the obtained DNN is not retrained after compression, leading to potentially sub-optimal solutions. Following this lead, the authors of~\cite{hou2016loss} have showed that training and compressing weights simultaneously can lead to better accuracy. In the same vein, in~\cite{courbariaux2015binaryconnect} and~\cite{soulie2016compression}, the authors propose to binarize the weights during the learning phase. As a result, the obtained DNN contains only weights whose values are $1$ or $-1$, while suffering from a very limited drop in accuracy compared to state-of-the-art solutions. These works have been improved later in~\cite{rastegari2016xnor}, where the authors proposed to add a scaling factor per layer and per kernel, as a mean to offer better diversity to binary networks, with almost no impact on memory usage or complexity. Other approaches have proposed to limit weights to three or more values ($-1$, $0$, and $1$)~\cite{lin2015neural,li2016ternary,zhou2016dorefa}. These approaches demonstrate that using slightly more bits to encode weights enable to improve accuracy by a significant amount, but they also require much more memory and other hardware components to compute non-binary operations. In recent works~\cite{courbariaux2016binarized,tang2017train}, authors have proposed to binarize both weights and activations in CNNs, resulting in potentially very efficient hardware implementations. However, these methods end up with a significant lower accuracy than state-of-the-art~ones.

Once a binary neural network has been trained, efficient implementations can advantageously benefit from simplified operations. For example, multiplications in binary neural networks can be replaced by simple low-cost multiplexers. Efficient solutions have been proposed in~\cite{andri2016yodann,ardakani2018convolutional}. However, even binary neural networks still require significant computational power and memory. These solutions also typically lead to a considerable latency, which may be an issue for some applications. In another line of work, authors have been aiming at reducing the number of trainable parameters in DNNs. In~\cite{han2015deep,ardakani2016sparsely}, the authors successfully apply pruning techniques to fully connected layers of DNNs. However, state-of-the-art CNNs are using more and more convolutional layers nowaday: in a typical modern architectures like ResNet18, about $99\%$ of the connections are in convolutional layers, and thus pruning connections only in fully connected layers has almost no impact on the overall complexity and memory usage of the architecture.

In this paper we propose to combine an efficient pruning technique, which can be effectively leveraged at implementation stage, with binary neural networks. We apply the proposed pruning technique on convolutional layers, resulting in very lightweight convolutions that can be implemented with simple multiplexers. The proposed method approaches state-of-art accuracy on the CIFAR10, CIFAR100 and SVHN dataset. We also propose a hardware implementation which uses very few resources and computational power. This implementation can compute more than one layer at a time and uses a simple multiplexer to perform convolutional operations. As such, it provides significantly smaller latency than existing counterparts.

The outline of the paper is as follows: in Section~\ref{proposedmethod} we describe the proposed method and describe experiments on the CIFAR10, CIFAR100 and SVHN dataset. In Section~\ref{HardwarePart} we present the proposed hardware implementation and show hardware implementation results. Section~\ref{conclusion} concludes.

\section{Proposed Method}
\label{proposedmethod}
In this section, we introduce a method to efficiently prune connections in convolutional layers. Note that pruning may have two different aims: a) to decrease the number of parameters to be trained in a given architecture, thus resulting in lesser chance of overfitting and b) to decrease the memory usage and complexity of a given architecture, so that it becomes lighter to implement in a budget-restricted configuration. If some author (e.g. ~\cite{ardakani2016sparsely}) argue they do both, we believe this is questionable as the reduction of the number of trainable weights they obtain on the one hand is balanced by the increasing complexity of identifying which connections are kept and which are lost in the process.

The proposed method has the double of interest of decreasing the number of parameters to be trained while keeping a simple deterministic way of identifying which connections are kept and which are disregarded. 

\subsection{Details of the Proposed Method}

Let us denote by $\mathbf{x}$ (resp. $\mathbf{y}$ or $\mathbf{w}$) the input (resp. output or kernel) tensor of a given convolutional layer. We index $\mathbf{x}$ (resp. $\mathbf{y}$) using three indices $i$, $j$, $k$ (resp. $\ell$), where $0\leq i < i_{\max}$ and $0\leq j < j_{\max}$ correspond to 2D coordinates and $0\leq k < k_{\max}$ (rsp. $0\leq \ell < \ell_{\max}$)ndexes a feature map. Similarly, we inde $\mathbf{w}$ using four indices: $0\leq \iota \leq \iota_{\max}$ and $0\leq \lambda \leq \lambda_{\max}$ correspond to 2D coordinates, and $k$ and $\ell$ are as introduced above. So, an element of the input tensor is written $x_{i,j,k}$, an element of the kernel tensor is written $w_{\iota,\lambda,k,l}$ and an element of the output tensor is written $y_{i,j,l}$.




The idea we propose consists of removing most of the connections in each slice $\mathbf{w}_{\cdot,\cdot,k,\ell}$ of the kernel tensor. The connections to be kept are chosen according to a deterministic rule agnostic of the initialization and of the training dataset. Namely, we choose to only keep the connections $w_{\iota,\lambda,k,\ell}$ for which 
\begin{equation} \iota + \lambda \iota_{\max} = k \pmod{\iota_{\max} \lambda_{\max}}.\label{equ:1} \end{equation}

When considering $3\times 3$ kernels for example, we remove 89\% of the connections in the convolutional layer. The reason for choosing this scheme is quite straightforward: we want diversity in the connections we keep to be sure our kernels do not simplify to a simple $1\times1$ convolution and still cover the initial kernel to its full extent (providing at least 9 feature maps are used).

We then perform the training on the remaining connections, disregarding the other ones. Using this method, the convolution of each slice of the kernel tensor is replaced by a simple multiplication. 

To further benefit from the reduced complexity of this pruning technique, we combine it with a weight binarization method. Here, we use Binary Connect (BC)~\cite{courbariaux2015binaryconnect}. Once remaining connections have been binarized, it is possible to replace the multiplication operation by a multiplexer.

\subsection{Results}

To evaluate the performance of our proposed method, we use the CIFAR10 vision benchmark made of tiny 32x32 images. We compare various modern CNN architectures such as Resnet~\cite{he2016deep}, Wide-Resnet~\cite{zagoruyko2016wide}, Densenet~\cite{huang2017densely}, and Mobilenet~\cite{sandler2018mobilenetv2}. Note that these architectures contain $1\times1$ and $3\times3$ convolutional kernels only. Thus we apply the proposed method on the $3\times3$ kernels.

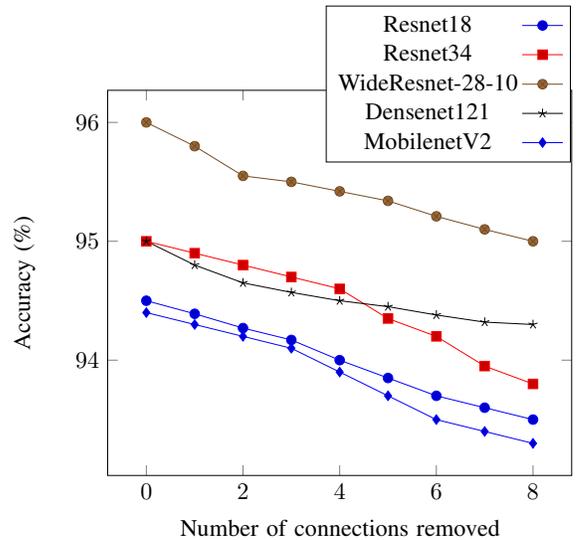
\begin{figure}[ht]
  \begin{center}

    \begin{tikzpicture}[scale=0.9]
\begin{axis}[xlabel=Number of connections removed,ylabel=Accuracy (\%),
legend style={at={(.47,1.22)},anchor=north west,},
legend entries={Resnet18, Resnet34, WideResnet-28-10, Densenet121, MobilenetV2 },
legend plot pos=right]


\addplot coordinates {(0,94.5)(1,94.39)(2,94.27)(3,94.17)(4,94)(5,93.85)(6,93.7)(7,93.6)(8,93.5)};
\addplot coordinates {(0,95)(1,94.9)(2,94.8)(3,94.7)(4,94.6)(5,94.35)(6,94.2)(7,93.95)(8,93.8)};
\addplot coordinates {(0,96)(1,95.8)(2,95.55)(3,95.5)(4,95.42)(5,95.34)(6,95.21)(7,95.1)(8,95)};
\addplot coordinates {(0,95)(1,94.8)(2,94.65)(3,94.57)(4,94.5)(5,94.45)(6,94.38)(7,94.32)(8,94.3)};
\addplot coordinates {(0,94.4)(1,94.3)(2,94.2)(3,94.1)(4,93.9)(5,93.7)(6,93.5)(7,93.4)(8,93.3)};
 \end{axis}
\end{tikzpicture}
  \end{center}
  \caption{Evolution of accuracy as function of number of connections removed per kernel slice.}
  \vspace{-0.45cm}
  \label{comp}
\end{figure}

As a first experiment, we aim at estimating the drop in performance caused by pruning connections. We thus randomly remove $m$ connections per kernel slice. Figure~\ref{comp} shows that the accuracy of the architecture is quite robust to this process, even when 8 out of the 9 connections in slices of $3\times3$ kernels are removed.

We then report in Table~\ref{table:table1} the obtained results using Equation~\eqref{equ:1} to remove kernels connections. Note that contrary to the previous experiment, removed connections are not chosen randomly anymore but according to a deterministic scheme. As a consequence, the positions of removed connections does not have to be stored. We compare the accuracy obtained using baseline architectures, pruned ones, binarized ones, and our proposed method mixing pruning and BC. Note that BC offers a 32 compression factor in terms of memory used, and our method roughly multiply this factor by 9, achieving an almost 300 factor compression in total. We also perform experiments on SVHN (resp. CIFAR100) on Resnet18 (resp. WideResnet-40-10) and obtain $97\%$/$96\%$ (resp. $80\%$/$77\%$) accuracy for Full-precision/pruning+BC.

\begin{table*}
\centering

    \caption{Comparison of accuracy between baseline architectures, pruned ones, binarized ones, and the proposed method on CIFAR10.}
    \centering
    {\renewcommand{\arraystretch}{1.3}%
    \begin{tabular} { | l || l | l | l | l | l |} 

  \cline{2-6}
    \multicolumn{1}{l|}{} & Resnet18 & Resnet34 & WideResnet-28-10 & Densenet121 & MobilenetV2\\
  
   \hline
  Full-precision & $94.5\%$ & $95\%$ & $96\%$ & $95\%$ & $94.4\%$ \\
  \hline
  Pruning   & $93.5\%$  & $93.8\%$ &  $95\%$ & $94.3\%$ & $93.3\%$\\
   \hline
   BC   & $93.31\%$  & $93.64\%$ &  $95.2\%$ & $94.5\%$ & $93\%$\\
   \hline
  Pruning + BC   & $91\%$  & $91.3\%$ &  $94\%$ & $93\%$ & $91\%$ \\
    
   \hline
    \end{tabular}} \quad
    \vspace{.3cm}
  
  \label{table:table1}
  \vspace{-.5cm}
  \end{table*}

\begin{table*}[h]
    \caption{FPGA results for the proposed architecture on vu13p (xcvu13p-figd2104-1-e).}
    \centering
    {\renewcommand{\arraystretch}{1.3}%
    \begin{tabular} { | l || l | l | l | l | l | l | l | l |} 
 \cline{2-9}
   \multicolumn{1}{l|}{} & P & LUT & FF & BRAMs & Frequency & Processing Latency & Processing outflow & Power\\
   \hline
  Conv$64-64$ & $16$ &$22424$ & $22424$ &$114$ &$240$MHz & $52\mu$ s & $19230$ images/s &$3.7$W\\
  \hline
  $4\times$Conv$64-64$ & $16$ & $89746$ & $75235$ &$456$ &$240$MHz & $208\mu$ s & $19230$ images/s & $6.5$W\\
  \hline
  $3\times$Conv$128-128$ & $32$ &$59780$ & $45024$ & $171$& $240$MHz & $154,8\mu$ s & $19379$ images/s & $4.8$W\\
  \hline
  $3\times$Conv$128-128$ & $64$ &$134090$ & $102552$ & $171$ & $240$MHz & $103,2\mu$ s & $29069$ images/s & $7.8$W\\
  \hline
  $3\times$Conv$256-256$ & $64$ &$74067$ & $52051$ & $87$&$250$MHz & $147,3\mu$ s & $20366$ images/s & $5.5$W\\
  \hline
  $3\times$Conv$256-256$ & $128$ &$ 154599$ & $102723$ & $87$&$218$MHz & $112,8\mu$ s & $26595$ images/s & $7.8$W\\
  \hline
  $3\times$Conv$512-512$ & $128$ &$132155$ & $52151$ &$45$ &$208$MHz & $177\mu$ & $16949$ images/s &$7.9$W\\
  \hline

    \end{tabular}} \quad
    \vspace{.3cm}
  
  \label{table:Results}
  \end{table*}

\section{Hardware Implementation}
\label{HardwarePart}

In this section, we first introduce the hardware architecture of the proposed method, its different components, and the way they are connected. Then, we present the hardware implementation of the proposed method, applied on ResNet18, on a Field Programmable Gate Array (FPGA).

\subsection{Hardware Architecture}
\label{hardwarearchitecture} 
 
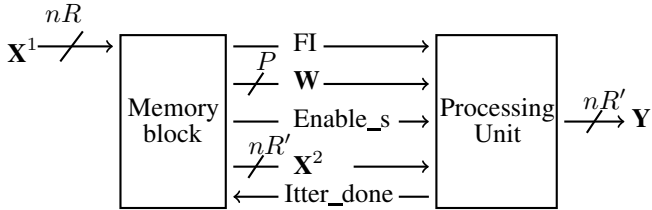
\begin{figure}[h]
    \begin{tikzpicture}[thick]
    \draw (0.6,-1.4)--(2,-1.4)--(2,.9)--(0.6,.9)--cycle;
\node[text width=3cm] at (2.2,-0.1) 
    {Memory};
\node[text width=3cm] at (2.4,-0.4) 
    {block};
\draw[->] (-.5,.75) -- (.5,.75) ;
\draw (.1,.95)--(-.2,.55);
\node[text width=3cm] at (1.1,1.2) 
    {$nR$};
\node[text width=3cm] at (.6,.7) 
    {$\textbf{X}^{1}$};
    
\draw (2.1,.75) -- (2.7,.75);
\node[text width=3cm] at (4.4,.8) 
    {FI};
\draw[->] (3.3,.75) -- (4.7,.75);

\draw (2.1,.25) -- (2.7,.25);
\draw (2.5,.4)--(2.3,.1);
\node[text width=3cm] at (3.9,.55) 
    {$P$};
\node[text width=3cm] at (4.4,.3) 
    {$\textbf{W}$};
\draw[->] (3.3,.25) -- (4.7,.25);

\draw (2.1,-.25) -- (2.7,-.25);

\node[text width=3cm] at (4.4,-.2) 
    {Enable\_s};
\draw[->] (4.3,-.25) -- (4.7,-.25);

\draw (2.1,-.85) -- (2.7,-.85);
\draw (2.5,-0.7)--(2.3,-1.);
\node[text width=3cm] at (3.8,-.55) 
    {$nR'$};
\node[text width=3cm] at (4.4,-.8) 
    {$\textbf{X}^{2}$};
\draw[->] (3.7,-.85) -- (4.7,-.85);

\def\x{4.2}
    
\draw (\x+0.6,-1.4)--(\x+2.2,-1.4)--(\x+2.2,.9)--(\x+0.6,.9)--cycle;
\node[text width=3cm] at (\x+2.15,-0.1) 
    {Processing};
\node[text width=3cm] at (\x+2.6,-0.4) 
    {Unit};

\draw[->] (\x+2.3,-.25) -- (\x+3.1,-.25);
\draw (\x+2.8,-0.1)--(\x+2.6,-.4);
\node[text width=3cm] at (\x+4.05,.05) 
    {$nR'$};
\node[text width=3cm] at (\x+4.7,-.2) 
    {$\textbf{Y}$};
    
\draw[<-] (2.1,-1.25) -- (2.7,-1.25);

\node[text width=3cm] at (4.28,-1.2) 
    {Itter\_done};
\draw (4.3,-1.25) -- (4.7,-1.25);

    \end{tikzpicture}
    \caption{Hardware architecture of a layer block.}
    \label{fig:layer}
\end{figure}

In Figure~\ref{fig:layer}, we depict the proposed hardware architecture for performing convolutions, which we name a ``layer block''. This architecture uses a simple low-cost multiplexer. In more details, a layer block is made of two sub-blocks: a memory one and a processing unit one.

The memory block contains two block RAMs (BRAMs) where content is encoded using $n$ bits fixed point. The first is used to store the computed feature maps. Once they are all computed, the content of the first BRAM is copied to the second one, so that it becomes the input of the next layers. At the same time, the computed feature maps of another image can be stored in the first BRAM. We thus obtain a pipeline architecture, in which all implemented layers work at the same time to speed up the classification process.

To avoid data overflow, we process each row of a slice of the input tensor $\mathbf{x}$ independently, and each slice of the kernel tensor independently. In more details, we copy from BRAM one to BRAM two a feature subvector $\mathbf{X}^2_{i,k}=\{x^2_{i,1,k},x^2_{i,2,k},\dots,x^2_{i,R',k}\}$ 
made of $R'$ values, instead of the whole feature vector $\mathbf{X}^1_{i,k}=\{x^1_{i,1,k},x^1_{i,2,k},\dots,x^1_{i,R,k}\}$  made of $R>R'$ values (cf.~Figure~\ref{fig:layer}). This is to account for the border effects (padding). To simplify notations, we replace $\mathbf{X}^1_{i,k}$ (resp. $\mathbf{X}^2_{i,k}$) by $\mathbf{X}^1$ (resp. $\mathbf{X}^2$) in the following.

\begin{figure*}[h]

\begin{tikzpicture}[thick, scale=0.9]

\draw[<-] (-.,-.5) -- (-.,.25) -- (.6,.25) ;
\draw[->] (-.,.25) -- (-.,2.65) -- (12.2,2.65) ;
\draw (.5,.45)--(.2,.05);
\node[text width=3cm] at (1.7,0.7) 
    {$nR'$};

\draw[->] (-1.1,.25) -- (-.5,.25) -- (-.5,-.5) ;
\draw (-.6,.45)--(-.9,.05);
\node[text width=3cm] at (0.6,0.7) 
    {$nR'$};
\node[text width=3cm] at (.15,0.3) 
    {$0$};
    
\draw (0.7,-0.2)--(2.1,-0.2)--(2.1,.9)--(0.7,.9)--cycle;
\node[text width=3cm] at (2.4,0.5) 
    {Register};
\node[text width=3cm] at (2.9,0.1) 
    {1};

\draw (-0.8,-0.7)--(.3,-0.7)--(.3,.-1.2)--(-0.8,-1.2)--cycle;
\node[text width=3cm] at (.95,-0.95) 
    {MUX};
\draw[->] (-1.5,-.95) -- (-.9,-.95);    
\node[text width=3cm] at (-.4,-0.95) 
    {FI};
\draw[->] (-.25,-1.3) -- (-.25,-2.05);

\draw (-.8,-2.15)--(.3,-2.15)--(.3,.-2.7)--(-.8,-2.7)--cycle;
\node[text width=3cm] at (1.1,-2.4) 
    {add};
\draw[->] (.4,-2.5) -- (1.3,-2.5) -- (1.3,-.3);
\draw (.6,-2.7)--(.9,-2.3);
\node[text width=3cm] at (2.1,-2.05) 
    {$nR'$};
\draw[<-] (-.25,-2.8) -- (-.25,-3.45);
\draw (-0.8,-3.6)--(.3,-3.6)--(.3,.-4.1)--(-0.8,-4.1)--cycle;
\node[text width=3cm] at (.95,-3.85) 
    {MUX};
\draw[->] (-1.5,-3.85) -- (-.9,-3.85);    
\node[text width=3cm] at (-.6,-3.85) 
    {$W_1$};   

\draw[<-] (-.,-4.3) -- (-.,-5.05) -- (.6,-5.05) ;
\draw (.5,-4.85)--(.2,.-5.25);
\node[text width=3cm] at (1.8,-4.6) 
    {$nR'$};

\draw[->] (-1.1,-5.05) -- (-.5,-5.05) -- (-.5,-4.3) ;
\draw (-.6,-4.85)--(-.9,-5.25);
\node[text width=3cm] at (.3,-4.6) 
    {$nR'$};
\node[text width=3cm] at (.,-5) 
    {$\textbf{X}^2$};
\node[text width=3cm] at (2.3,-5) 
    {$-\textbf{X}^2$};
\def\x{5}
\draw[<-] (-.+\x,-.5) -- (-.+\x,.25) -- (.6+\x,.25) ;
\draw[->] (-.+\x,.25) -- (-.+\x,2.35) -- (12.2,2.35) ;
\draw (.5+\x,.45)--(.2+\x,.05);
\node[text width=3cm] at (1.7+\x,0.7) 
    {$nR'$};

\draw[->] (-1.1+\x,.25) -- (-.5+\x,.25) -- (-.5+\x,-.5) ;
\draw (-.6+\x,.45)--(-.9+\x,.05);
\node[text width=3cm] at (.5+\x,0.7) 
    {$nR'$};
\node[text width=3cm] at (.15+\x,0.3) 
    {$0$};
    
\draw (0.7+\x,-0.2)--(2.1+\x,-0.2)--(2.1+\x,.9)--(0.7+\x,.9)--cycle;
\node[text width=3cm] at (2.4+\x,0.5) 
    {Register};
\node[text width=3cm] at (2.9+\x,0.1) 
    {2};

\draw (-0.8+\x,-0.7)--(.3+\x,-0.7)--(.3+\x,.-1.2)--(-0.8+\x,-1.2)--cycle;
\node[text width=3cm] at (.95+\x,-0.95) 
    {MUX};
\draw[->] (-1.5+\x,-.95) -- (-.9+\x,-.95);    
\node[text width=3cm] at (-.4+\x,-0.95) 
    {FI};
\draw[->] (-.25+\x,-1.3) -- (-.25+\x,-2.05);

\draw (-.8+\x,-2.15)--(.3+\x,-2.15)--(.3+\x,.-2.7)--(-.8+\x,-2.7)--cycle;
\node[text width=3cm] at (1.1+\x,-2.4) 
    {add};
\draw[->] (.4+\x,-2.5) -- (1.3+\x,-2.5) -- (1.3+\x,-.3);
\draw (.6+\x,-2.7)--(.9+\x,-2.3);
\node[text width=3cm] at (2.1+\x,-2.05) 
    {$nR'$};
\draw[<-] (-.25+\x,-2.8) -- (-.25+\x,-3.45);
\draw (-0.8+\x,-3.6)--(.3+\x,-3.6)--(.3+\x,.-4.1)--(-0.8+\x,-4.1)--cycle;
\node[text width=3cm] at (.95+\x,-3.85) 
    {MUX};
\draw[->] (-1.5+\x,-3.85) -- (-.9+\x,-3.85);    
\node[text width=3cm] at (-.6+\x,-3.85) 
    {$W_2$};

\draw[<-] (-.+\x,-4.3) -- (-.+\x,-5.05) -- (.6+\x,-5.05) ;
\draw (.5+\x,-4.85)--(.2+\x,.-5.25);
\node[text width=3cm] at (1.8+\x,-4.6) 
    {$nR'$};

\draw[->] (-1.1+\x,-5.05) -- (-.5+\x,-5.05) -- (-.5+\x,-4.3) ;
\draw (-.6+\x,-4.85)--(-.9+\x,-5.25);
\node[text width=3cm] at (.3+\x,-4.6) 
    {$nR'$};
\node[text width=3cm] at (.+\x,-5) 
    {$\textbf{X}^2$};
\node[text width=3cm] at (2.3+\x,-5) 
    {$-\textbf{X}^2$};

\draw[dashed] (7,-2.2) -- (10,-2.2) ;

    
\def\x2{11.5}
\draw[<-] (-.+\x2,-.5) -- (-.+\x2,.25) -- (.6+\x2,.25) ;

\draw[dashed] (.35+\x2,1.9)--(.35+\x2,2.3);
\draw[->] (-.+\x2,.25) -- (-.+\x2,1.75) -- (.7+\x2,1.75) ;
\draw (.5+\x2,.45)--(.2+\x2,.05);
\node[text width=3cm] at (1.7+\x2,0.7) 
    {$nR'$};

\draw[->] (-1.1+\x2,.25) -- (-.5+\x2,.25) -- (-.5+\x2,-.5) ;
\draw (-.6+\x2,.45)--(-.9+\x2,.05);
\node[text width=3cm] at (.5+\x2,0.7) 
    {$nR'$};
\node[text width=3cm] at (.15+\x2,0.3) 
    {$0$};
    
\draw (0.7+\x2,-0.2)--(2.1+\x2,-0.2)--(2.1+\x2,.9)--(0.7+\x2,.9)--cycle;
\node[text width=3cm] at (2.4+\x2,0.5) 
    {Register};
\node[text width=3cm] at (2.9+\x2,0.1) 
    {$P$};

\draw (-0.8+\x2,-0.7)--(.3+\x2,-0.7)--(.3+\x2,.-1.2)--(-0.8+\x2,-1.2)--cycle;
\node[text width=3cm] at (.95+\x2,-0.95) 
    {MUX};
\draw[->] (-1.5+\x2,-.95) -- (-.9+\x2,-.95);    
\node[text width=3cm] at (-.4+\x2,-0.95) 
    {FI};
\draw[->] (-.25+\x2,-1.3) -- (-.25+\x2,-2.05);

\draw (-.8+\x2,-2.15)--(.3+\x2,-2.15)--(.3+\x2,.-2.7)--(-.8+\x2,-2.7)--cycle;
\node[text width=3cm] at (1.1+\x2,-2.4) 
    {add};
\draw[->] (.4+\x2,-2.5) -- (1.3+\x2,-2.5) -- (1.3+\x2,-.3);
\draw (.6+\x2,-2.7)--(.9+\x2,-2.3);
\node[text width=3cm] at (2.1+\x2,-2.05) 
    {$nR'$};
\draw[<-] (-.25+\x2,-2.8) -- (-.25+\x2,-3.45);
\draw (-0.8+\x2,-3.6)--(.3+\x2,-3.6)--(.3+\x2,.-4.1)--(-0.8+\x2,-4.1)--cycle;
\node[text width=3cm] at (.95+\x2,-3.85) 
    {MUX};
\draw[->] (-1.5+\x2,-3.85) -- (-.9+\x2,-3.85);    
\node[text width=3cm] at (-.6+\x2,-3.85) 
    {$W_P$};

\draw[<-] (-.+\x2,-4.3) -- (-.+\x2,-5.05) -- (.6+\x2,-5.05) ;
\draw (.5+\x2,-4.85)--(.2+\x2,.-5.25);
\node[text width=3cm] at (1.8+\x2,-4.6) 
    {$nR'$};

\draw[->] (-1.1+\x2,-5.05) -- (-.5+\x2,-5.05) -- (-.5+\x2,-4.3) ;
\draw (-.6+\x2,-4.85)--(-.9+\x2,-5.25);
\node[text width=3cm] at (.3+\x2,-4.6) 
    {$nR'$};
\node[text width=3cm] at (.+\x2,-5) 
    {$\textbf{X}^2$};
\node[text width=3cm] at (2.3+\x2,-5) 
    {$-\textbf{X}^2$};

\draw (0.9+\x2,4.6)--(2.3+\x2,4.6)--(2.3+\x2,3.8)--(0.9+\x2,3.8)--cycle;
\node[text width=3cm] at (2.6+\x2,4.2) 
    {Counter};
\draw[->] (1.6+\x2,3.6)--(1.6+\x2,3); 
\draw[->] (.+\x2,4)--(0.8+\x2,4); 
\node[text width=3cm] at (.+\x2,4.1) 
    {Enable\_s};

\draw[<-] (.+\x2,4.3)--(0.8+\x2,4.3); 
\node[text width=3cm] at (.+\x2-.15,4.4) 
    {Itter\_done};
\draw (0.8+\x2,2.8)--(2.3+\x2,2.8)--(2.3+\x2,1.6)--(0.8+\x2,1.6)--cycle;
\node[text width=3cm] at (2.5+\x2,2.2) 
    {DEMUX};

\draw[->] (2.5+\x2,2.2) -- (3+\x2,2.2) ;

\draw (3.1+\x2,2.5)--(4+\x2,2.5)--(4+\x2,1.9)--(3.1+\x2,1.9)--cycle;
\node[text width=3cm] at (4.8+\x2,2.2) 
    {Relu};
\draw[->] (4.1+\x2,2.2) -- (4.8+\x2,2.2) ;
\draw (4.5+\x2,2.35) -- (4.3+\x2,2.05) ;
\node[text width=3cm] at (\x2+5.8,2.6) 
    {$nR'$};
\node[text width=3cm] at (\x2+6.6,2.2) 
    {\textbf{Y}};

\end{tikzpicture}
\caption{Hardware architecture of a processing unit block.}
\label{fig: process_unit}
\end{figure*}
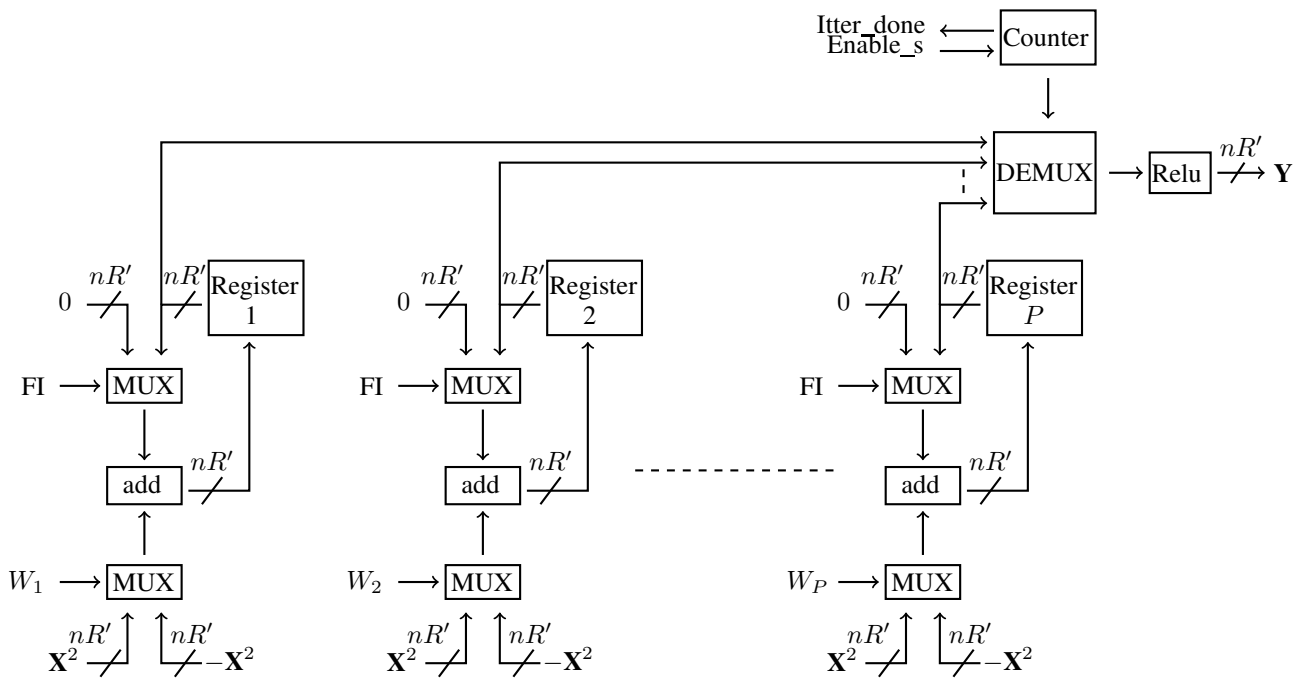

\begin{figure}[H]
 \begin{center}
     \begin{tikzpicture}[thick]
     \draw (0.6,-1.2)--(8.4,-1.2)--(8.4,-.6)--(0.6,-.6)--cycle;
     \node[text width=3cm] at (2.25,-.95) 
     {$x^1_{i,1,k}$};
     \node[text width=3cm] at (3.25,-.95) 
     {$x^1_{i,2,k}$};
     \node[text width=3cm] at (4.25,-.95) 
     {$x^1_{i,3,k}$};
     \node[text width=3cm] at (5.25,-.95) 
     {$x^1_{i,4,k}$};
     \draw[dashed] (4.7,-0.9)--(5.7,-.9);
     \node[text width=3cm] at (7.45,-.95)
     {$x^1_{i,R-1,k}$};
     \node[text width=3cm] at (8.95,-.95) 
     {$x^1_{i,R,k}$};
    \draw[->](.9,-1.1)--(.9,-2);
     \draw[->](1.9,-1.1)--(1.9,-2);
     \draw (0.65,-2.1)--(2.15,-2.1)--(2.15,-3)--(0.65,-3)--cycle;
     \node[text width=3cm] at (2.47,-2.55) 
     {MUX};
     \draw[->](1.37,-3.1)--(1.37,-4);

     \draw[->](2.9,-1.1)--(2.9,-2);
     \draw[->](3.9,-1.1)--(3.9,-2);
     \draw (2.65,-2.1)--(4.15,-2.1)--(4.15,-3)--(2.65,-3)--cycle;
     \node[text width=3cm] at (4.47,-2.55) 
     {MUX};
     \draw[->](3.37,-3.1)--(3.37,-4);

     \def\x{3.7}
     \draw[->](\x+2.9,-1.1)--(\x+2.9,-2);
     \draw[->](\x+3.9,-1.1)--(\x+3.9,-2);
     \draw (\x+2.65,-2.1)--(\x+4.15,-2.1)--(\x+4.15,-3)--(\x+2.65,-3)--cycle;
     \node[text width=3cm] at (\x+4.47,-2.55) 
     {MUX};
     \draw[->](\x+3.37,-3.1)--(\x+3.37,-4);

     \draw (.67,-4.7)--(7.45,-4.7)--(7.45,-4.1)--(.67,-4.1)--cycle;
     \node[text width=3cm] at (2.3,-4.45) 
     {$x^2_{i,1,\ell}$};
     \node[text width=3cm] at (4.3,-4.45) 
     {$x^2_{i,2,\ell}$};
     \draw[dashed](3.9,-4.4)--(6,-4.4);
     \node[text width=3cm] at (\x+4.2,-4.45) 
     {$x^2_{i,R',\ell}$};
     \end{tikzpicture}
     \end{center}
 \caption{Hardware architecture to copy the first BRAM contents to the second BRAM, when stride value is $2$. }
     \label{fig:my_label}
 \end{figure}
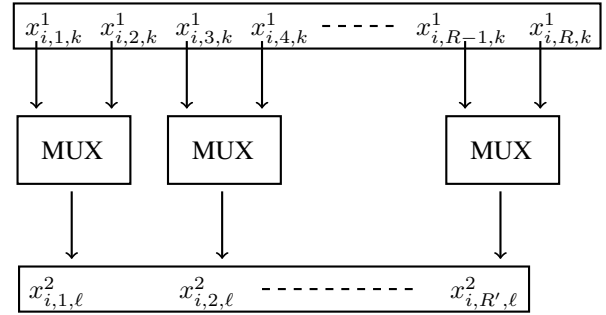

To compute a convolutional operation, kernels move along feature maps with a step which is called stride in CNNs. In a typical case in which stride value is $1$, $\textbf{X}^2$ represents either the first $R'$ values, the middle $R'$ ones or the last $R'$ ones, where $R' = R - 2$, depending on the position of the nonzero kernel value. When stride value is $2$ (c.f. Figure~\ref{fig:my_label}), only half of the values are copied from $\textbf{X}^1$ to $\textbf{X}^2$ by selecting either the odd or even values of $j$ in $x^1_{i,j,k}$ using multiplexers. This process can be generalised to any stride value other than 1 or 2.


The processing unit uses $\textbf{X}^{2}$ and a vector $\textbf{W}$ made of $P$ values coded on $1$ bit each corresponding to weights in the convolution kernel. It thus computes in parallel $P$ feature vectors (cf.~Figure~\ref{fig: process_unit}). The First-Input signal (FI) is set to $1$ when the first feature vector is read from the second BRAM to initialise registers by $0$. To compute each feature vector $p$ where $1 \leq p \leq P$, we use the corresponding $W_p$ to add either $\textbf{X}^2$ or $\textbf{-X}^2$ to the content of register $p$. Once all input feature vectors have been read from the second BRAM of memory block, the signal \texttt{Enable\_s} is set to $1$, and the content of registers is written one by one into the first BRAM of the memory block of the next layer. At the end of this process, the \texttt{Itter\_done} signal is set to $1$ in the processing unit block, so new data can be read from the memory block to process other feature vectors. 

To achieve the computation associated with the layer block described in Figure~\ref{fig:layer}, $k_{max}j_{max}$ clock cycles (CCs) are required to copy all contents from the first BRAM to the second one, $j_{max}k_{max}\ell_{max}/P$ CCs to compute all output feature vectors of one layer, and $j_{max}\ell_{max}$ CCs to write all computed feature vectors into the memory block of the next layer. Thus the total number of CCs required is:

\begin{equation}
CCs = j_{max}k_{max} + \frac{j_{max}k_{max}\ell_{max}}{P} +j_{max}\ell_{max}    
\label{equation:CCs1}
\end{equation}

This should be compared to~\cite{ardakani2018convolutional}, where the number of clock cycles becomes:

\begin{equation}
CCs = \frac{3j_{max}^2k_{max}\ell_{max}}{P}    
\label{equation:CCs2}
\end{equation}

We observe that the proposed architecture is $3j_{max}$ faster than~\cite{ardakani2018convolutional}, which can be significant when $j_{max}$ is big. For instance with the CIFAR10 dataset, at the input layer of a CNN $j_{max}=32$, and thus the proposed method is $96$ times faster. In addition it is a pipeline architecture, so it can be $3Lj_{max}$ faster where $L$ is the total number of layer blocks that fit in an FPGA.

Note that in the proposed architecture, $P$ should be lower or equal to  $\ell_{max}$, otherwise reaching full parallelism would require to read more than one vector $\textbf{X}^2$, and as such would also require more BRAMs, resulting in a more complex architecture.

\subsection{Hardware Results}

We implemented one/few layers of Resnet18 on Xilinx Ultra Scale Vu13p (xcvu13p-figd2104-1-e) FPGA. The implemented layers are arranged in a pipeline, and their functionality has been verified comparing the output of each layer block with the ones obtained by software simulation over a batch of examples. Table~\ref{table:Results} shows the required resources to implement one/few layers of Resnet18 trained on CIFAR10 dataset for different values of $P$. It also shows that the obtained architecture obtain a low processing latency to compute a valid output of one layer. Moreover, this processing latency increases when processing more than one layer, but processing outflow is 
maintained thanks to the pipeline design.

\section{Conclusion}
\label{conclusion}

In this paper, we proposed to extend pruning techniques to convolutional layers of DNNs. We introduced a deterministic pruning scheme that can be taken advantage of in implementations. We combined pruning with weight binarization to reduce both complexity and memory usage and showed the resulting neural network is still able to reach very high accuracy.

We implemented the proposed scheme using a low cost hardware architecture in which complex convolution operations are replaced by simple multiplexers. As a result, we were able to implement a considerable part of some complex CNNs such as Resnet18. Moreover, the architecture only consumes a few watts, making it a good solution for embedded applications. Future work will extend this method to all kernel shapes, and propose a low cost hardware architecture to handle other challenging vision datasets such as ImageNet. 
\bibliographystyle{IEEEtran}
\bibliography{ref}

\begin{thebibliography}{10}
\providecommand{\url}[1]{#1}
\csname url@samestyle\endcsname
\providecommand{\newblock}{\relax}
\providecommand{\bibinfo}[2]{#2}
\providecommand{\BIBentrySTDinterwordspacing}{\spaceskip=0pt\relax}
\providecommand{\BIBentryALTinterwordstretchfactor}{4}
\providecommand{\BIBentryALTinterwordspacing}{\spaceskip=\fontdimen2\font plus
\BIBentryALTinterwordstretchfactor\fontdimen3\font minus
  \fontdimen4\font\relax}
\providecommand{\BIBforeignlanguage}[2]{{%
\expandafter\ifx\csname l@#1\endcsname\relax
\typeout{** WARNING: IEEEtran.bst: No hyphenation pattern has been}%
\typeout{** loaded for the language `#1'. Using the pattern for}%
\typeout{** the default language instead.}%
\else
\language=\csname l@#1\endcsname
\fi
#2}}
\providecommand{\BIBdecl}{\relax}
\BIBdecl

\bibitem{lecun1998gradient}
Y.~LeCun, L.~Bottou, Y.~Bengio, and P.~Haffner, ``Gradient-based learning
  applied to document recognition,'' \emph{Proceedings of the IEEE}, vol.~86,
  no.~11, pp. 2278--2324, 1998.

\bibitem{iandola2016squeezenet}
F.~N. Iandola, S.~Han, M.~W. Moskewicz, K.~Ashraf, W.~J. Dally, and K.~Keutzer,
  ``Squeezenet: Alexnet-level accuracy with 50x fewer parameters and $<$ 0.5 mb
  model size,'' \emph{arXiv preprint arXiv:1602.07360}, 2016.

\bibitem{DBLP:journals/corr/SimonyanZ14a}
\BIBentryALTinterwordspacing
K.~Simonyan and A.~Zisserman, ``Very deep convolutional networks for
  large-scale image recognition,'' \emph{CoRR}, vol. abs/1409.1556, 2014.
  [Online]. Available: \url{http://arxiv.org/abs/1409.1556}
\BIBentrySTDinterwordspacing

\bibitem{DBLP:journals/corr/Graham14a}
\BIBentryALTinterwordspacing
B.~Graham, ``Fractional max-pooling,'' \emph{CoRR}, vol. abs/1412.6071, 2014.
  [Online]. Available: \url{http://arxiv.org/abs/1412.6071}
\BIBentrySTDinterwordspacing

\bibitem{szegedy2015rethinking}
C.~Szegedy, V.~Vanhoucke, S.~Ioffe, J.~Shlens, and Z.~Wojna, ``Rethinking the
  inception architecture for computer vision,'' \emph{arXiv preprint
  arXiv:1512.00567}, 2015.

\bibitem{lecun2015deep}
Y.~LeCun, Y.~Bengio, and G.~Hinton, ``Deep learning,'' \emph{nature}, vol. 521,
  no. 7553, p. 436, 2015.

\bibitem{han2015deep}
S.~Han, H.~Mao, and W.~J. Dally, ``Deep compression: Compressing deep neural
  networks with pruning, trained quantization and huffman coding,'' \emph{arXiv
  preprint arXiv:1510.00149}, 2015.

\bibitem{kim2015compression}
Y.-D. Kim, E.~Park, S.~Yoo, T.~Choi, L.~Yang, and D.~Shin, ``Compression of
  deep convolutional neural networks for fast and low power mobile
  applications,'' \emph{arXiv preprint arXiv:1511.06530}, 2015.

\bibitem{hou2016loss}
L.~Hou, Q.~Yao, and J.~T. Kwok, ``Loss-aware binarization of deep networks,''
  \emph{arXiv preprint arXiv:1611.01600}, 2016.

\bibitem{courbariaux2015binaryconnect}
M.~Courbariaux, Y.~Bengio, and J.-P. David, ``Binaryconnect: Training deep
  neural networks with binary weights during propagations,'' in \emph{Advances
  in neural information processing systems}, 2015, pp. 3123--3131.

\bibitem{soulie2016compression}
G.~Souli{\'e}, V.~Gripon, and M.~Robert, ``Compression of deep neural networks
  on the fly,'' in \emph{International Conference on Artificial Neural
  Networks}.\hskip 1em plus 0.5em minus 0.4em\relax Springer, 2016, pp.
  153--160.

\bibitem{rastegari2016xnor}
M.~Rastegari, V.~Ordonez, J.~Redmon, and A.~Farhadi, ``Xnor-net: Imagenet
  classification using binary convolutional neural networks,'' in
  \emph{European Conference on Computer Vision}.\hskip 1em plus 0.5em minus
  0.4em\relax Springer, 2016, pp. 525--542.

\bibitem{lin2015neural}
Z.~Lin, M.~Courbariaux, R.~Memisevic, and Y.~Bengio, ``Neural networks with few
  multiplications,'' \emph{arXiv preprint arXiv:1510.03009}, 2015.

\bibitem{li2016ternary}
F.~Li, B.~Zhang, and B.~Liu, ``Ternary weight networks,'' \emph{arXiv preprint
  arXiv:1605.04711}, 2016.

\bibitem{zhou2016dorefa}
S.~Zhou, Y.~Wu, Z.~Ni, X.~Zhou, H.~Wen, and Y.~Zou, ``Dorefa-net: Training low
  bitwidth convolutional neural networks with low bitwidth gradients,''
  \emph{arXiv preprint arXiv:1606.06160}, 2016.

\bibitem{courbariaux2016binarized}
M.~Courbariaux, I.~Hubara, D.~Soudry, R.~El-Yaniv, and Y.~Bengio, ``Binarized
  neural networks: Training deep neural networks with weights and activations
  constrained to+ 1 or-1,'' \emph{arXiv preprint arXiv:1602.02830}, 2016.

\bibitem{tang2017train}
W.~Tang, G.~Hua, and L.~Wang, ``How to train a compact binary neural network
  with high accuracy?'' in \emph{AAAI}, 2017, pp. 2625--2631.

\bibitem{andri2016yodann}
R.~Andri, L.~Cavigelli, D.~Rossi, and L.~Benini, ``Yodann: An ultra-low power
  convolutional neural network accelerator based on binary weights.'' in
  \emph{ISVLSI}, 2016, pp. 236--241.

\bibitem{ardakani2018convolutional}
A.~Ardakani, C.~Condo, and W.~J. Gross, ``A convolutional accelerator for
  neural networks with binary weights,'' in \emph{Circuits and Systems (ISCAS),
  2018 IEEE International Symposium on}.\hskip 1em plus 0.5em minus 0.4em\relax
  IEEE, 2018, pp. 1--5.

\bibitem{ardakani2016sparsely}
------, ``Sparsely-connected neural networks: towards efficient vlsi
  implementation of deep neural networks,'' \emph{arXiv preprint
  arXiv:1611.01427}, 2016.

\bibitem{he2016deep}
K.~He, X.~Zhang, S.~Ren, and J.~Sun, ``Deep residual learning for image
  recognition,'' in \emph{Proceedings of the IEEE conference on computer vision
  and pattern recognition}, 2016, pp. 770--778.

\bibitem{zagoruyko2016wide}
S.~Zagoruyko and N.~Komodakis, ``Wide residual networks,'' \emph{arXiv preprint
  arXiv:1605.07146}, 2016.

\bibitem{huang2017densely}
G.~Huang, Z.~Liu, L.~Van Der~Maaten, and K.~Q. Weinberger, ``Densely connected
  convolutional networks.'' in \emph{CVPR}, vol.~1, no.~2, 2017, p.~3.

\bibitem{sandler2018mobilenetv2}
M.~Sandler, A.~Howard, M.~Zhu, A.~Zhmoginov, and L.-C. Chen, ``Mobilenetv2:
  Inverted residuals and linear bottlenecks,'' in \emph{Proceedings of the IEEE
  Conference on Computer Vision and Pattern Recognition}, 2018, pp. 4510--4520.

\end{thebibliography}

\end{document}